\pdfoutput=1
\documentclass[11pt]{article}

\usepackage[]{acl}

\usepackage{times}
\usepackage{latexsym}

\usepackage[T1]{fontenc}
\usepackage[utf8]{inputenc}

\usepackage{microtype}

\usepackage{amssymb}
\usepackage{amsmath} 
\usepackage{booktabs}
\usepackage{enumerate}
\usepackage{graphicx}
\usepackage{subfigure}
\usepackage{xspace}
\usepackage{float}
\usepackage{bbm}
\usepackage{bm}
\usepackage{multirow}
\usepackage{booktabs}
\usepackage{color}
\usepackage{framed}
\usepackage{stfloats}
\usepackage{iitem}
\usepackage[T1]{fontenc}
\definecolor{shadecolor}{RGB}{180,180,180}
\usepackage{url}
\newcommand{\paratitle}[1]{\vspace{1.5ex}\noindent\textbf{#1}}
\newcommand{\ie}{\emph{i.e.,}\xspace}

\newcommand{\eg}{\emph{e.g.,}\xspace}

\newcommand{\ignore}[1]{}

\usepackage{algorithm}

\usepackage{algpseudocode}
\usepackage{amsthm}
\usepackage{amsmath}
\usepackage{tikz}

\usepackage{amsmath}

\newcommand\Matrix{\mathbf}            % \matrix is defined in LaTeX2e kernel
\newcommand\Tensor{\mathcal}

\usepackage{colortbl}
\usepackage{color, xcolor}
\title{Parameter-Efficient Mixture-of-Experts Architecture \\ for Pre-trained Language Models}

\author{
	Ze-Feng Gao$^{1,4,5}$\thanks{$\ $ Authors contributed equally.}\ \ ,
	Peiyu Liu$^{1,4*}$\ ,
	Wayne Xin Zhao$^{1,4}$\thanks{$\ $ Corresponding author.}\ \ ,
	\textbf{Zhong-Yi Lu}$^{2}$\and
	\textbf{Ji-Rong Wen}$^{1,3,4}$
	\\
	$^1$Gaoling School of Artificial Intelligence, Renmin University of China\\
	$^2$Department of Physics, Renmin University of China\\
	$^3$ School of Information, Renmin University of China\\
	$^4$Beijing Key Laboratory of Big Data Management and Analysis Methods\\
	$^5$Beijing Academy of Artificial Intelligence, Beijing, 100084, China\\
	{\{zfgao,liupeiyustu,zlu,jrwen\}@ruc.edu.cn, batmanfly@gmail.com}\\ 
}

\begin{document}
\maketitle

\begin{abstract}
Recently,  Mixture-of-Experts~(short as MoE) architecture has achieved  remarkable success in increasing the model capacity of large-scale language models. 
However, MoE requires incorporating significantly more parameters than the base model being extended. 
In this paper, we propose building a parameter-efficient MoE architecture by sharing information among experts.
We adopt matrix product operator~(MPO, a tensor decomposition from quantum many-body physics) to reconstruct the parameter matrix in the expert layer and increase model capacity for pre-trained language models by sharing parameters of the central tensor~(containing the core information) among different experts while enabling the specificity through the auxiliary tensors~(complementing the central tensor) of different experts.
To address the unbalanced optimization issue, we further design the gradient mask strategy for the MPO-based MoE architecture.
Extensive experiments based on T5 and GPT-2 show improved performance and efficiency of the pre-trained language model~(27.2x reduction in total parameters for the superior model performance, compared with the Switch Transformers). 
Our code is publicly available at \url{https://github.com/RUCAIBox/MPOE}.
% We also demonstrate an improvement in  multi-task learning across a variety of tasks.
% \textcolor{blue}{We also observe benefits of sharing information across a variety of tasks in multi-task learning.}
% We also obtain potential benefits in multi-task learning due to sharing information across a variety of tasks .
% We also obtain performance gains benefits in multi-task learning due to sharing information across a variety of tasks .
% \textcolor{blue}{Performance improvements of experiments in multi-task learning also demonstrate the benefits of sharing information across different tasks.}
% \textcolor{blue}{Additionally, performance gains in GLUE benchmark show that sharing parameters of central tensors improve multi-task learning.}
% \textcolor{blue}{Additionally, our approach achieves  performance gains in GLUE benchmark, suggesting that sharing parameters of central tensors can also improve multi-task learning.}
% \textcolor{blue}{Additionally, our approach achieves performance gains in few-shot domain generalization across various tasks.}

\end{abstract}
\section{Introduction}

Large-scale pre-trained language models~(PLMs), such as BERT~\cite{devlin2018bert} and T5~\cite{multitask2020raffel}, have become the de facto standard in natural language processing~(NLP). By involving a huge number of parameters pre-trained on the general-purpose corpus, PLMs can achieve excellent performance in many NLP tasks. 
In order to increase the model capacity, a promising direction is to explore the scaling properties  with the mixture-of-experts~(MoE) paradigm~\cite{jacobs1991adaptive,shazeer2017outrageously} for  developing more powerful PLMs.
By incorporating multiple expert networks, MoE schedules the learning of data samples through a routing component that is usually implemented by some gating function, which increases model capacity without a proportional increase in computation costs.
Despite the effectiveness, it has been shown that the MoE architecture is parameter inefficient~\cite{zuo2021thor},  considering the yielded improvement \emph{w.r.t.} the involved costs. 
Most of the existing studies~\cite{yang2022sparse,roller2021hash,lewis2021base} attribute this issue to the unbalanced load of experts, focusing on improving the routing strategies. 

However, an important question  about the MoE architecture has been neglected in previous studies: whether the increased parameters from the experts are all necessary to increase the model capacity. As different experts from an MoE network are often trained with correlated data samples~(\eg sample correlation from training data), it is likely to lead to parameter redundancy across experts.
% ~指出在不同专家之间具有荣有冗余性信息，因此在使用稀疏模型部署的时候，采用蒸馏或者裁减掉冗余专家的方法来改善这个问题。
% Indeed, \emph{expert redundancy} has been identified in  existing studies~\citep{fedus2021switch,kim2021scalable}, where they distilled or pruned  redundant experts with sparse models.
Indeed, \emph{expert redundancy} has been identified in  existing studies, where~\citet{fedus2021switch} distills sparse MoE models into dense models and~\citet{kim2021scalable} prunes experts to compress MoE models.
%Existing studies~\citep{fedus2021switch,kim2021scalable} pointed out that there is redundant information among different experts 
% redundant experts \textcolor{blue}{with sparse models}.
% 我们也利用了MMD对不同专家的冗余性做了经验性分析，结果表明给定相同输入的专家所输出的特征具有相同的分布。
% We also used Maximum Mean Discrepancy~(MMD)~\cite{2007mmd} to empirically analyze the redundancy of different experts, and the results show that the output feature by different experts given the same inputs have the same distribution~(See Appendix.E).
This finding motivates us to develop a parameter-efficient MoE architecture by reducing its parameter redundancy. 
Intuitively, a straightforward approach is to share a certain  proportion of parameters among experts. However, it is difficult to identify and optimize the key parameters that encode the shared information across experts, since expert networks typically consist of dense matrices.

To address this issue, we propose a novel parameter sharing approach inspired by the matrix product operators~(MPO) decomposition from quantum many-body physics~\cite{gao2020compressing}, which decomposes a matrix into a sequential product of local tensors~(either \emph{central} or \emph{auxiliary} tensors). 
Unlike other matrix decomposition methods, MPO can effectively reorganize and aggregate important information of the original matrix into the \emph{central tensor}. 
The auxiliary tensors, on the other hand, serve to complement the central tensor for recovering the original matrix~\cite{DBLP:conf/acl/LiuGZXLW20}.
In the setting of MoE, considering the small parameter variations among experts, we speculate that the central tensors of different experts~(with MPO decomposition for each expert) are likely to be very similar. If the central tensors could be shared for all expert networks, we would significantly reduce the parameters of the MoE architecture.

To this end, we propose a novel \underline{MPO}-based parameter-efficient Mo\underline{E} architecture, called \textbf{MPOE}.
Based on classic MoE architecture~\cite{shazeer2017outrageously}, 
our approach introduces a major extension allowing experts to share a global central tensor while keeping expert-specific auxiliary tensors. %, with the MPO decomposition.
%We obtain the  central and auxiliary tensors with the MPO decomposition. 
In our setting, the parameter matrix of in a single expert is formed by the product of the globally shared central tensor and the corresponding auxiliary tensors.
%For each expert, we can reconstruct its information via the product between the shared central tensor and local auxiliary tensors. 
Since the central tensor contains most of the parameters from an MPO decomposition, our MPOE approach can significantly reduce the parameters of the original MoE architecture. Another major merit of MPO is that auxiliary tensors are closely entangled with the central tensor~\cite{pirvu2010matrix}, and it is theoretically guaranteed that any change from auxiliary tensors can be propagated to the central tensor. That is to say, though a large proportion of parameters are shared, local auxiliary tensors still enable the experts to capture specific variations or differences according to routing data samples. 
% \textcolor{blue}{However, directly optimizing the MPOE architecture is likely to lead to the unbalanced optimization issue~\cite{fire2019over,stiennon2020learning}, since the central tensors are updated more frequently than auxiliary tensors during fine-tuning.}
% 由于优化次数过多而
However, directly optimizing the MPOE architecture is likely to lead to an \emph{unbalanced optimization} issue, where the central tensors are updated more frequently than auxiliary tensors during fine-tuning.
Therefore, we further propose a gradient mask strategy that masks the central tensor gradient to effectively alleviate the unbalanced optimization issue. 

To the best of our knowledge, this is the first attempt to reduce the parameter redundancy of the MoE architecture with structural matrix decomposition.  We conduct extensive experiments to evaluate the effectiveness of the MPOE architecture on two representatives PLMs, T5 and GPT. 
Experiments have demonstrated the effectiveness of our approach in increasing model capacity~(27.2x fewer parameters for the superior model performance, compared with several competitive MoE-enhanced PLMs. 
%\textcolor{blue}{compared with the Switch Transformer~\citep{fedus2021switch}, MoE Transformer~\cite{shazeer2017outrageously} and MoEfication~\cite{zhang2021moefication}).}

\section{Preliminary}\label{sec-preliminary}
% In this section, we first describe the MoE, and then introduce the MPO decomposition, which serves as a basic technique to implement our approach.

\subsection{Mixture-of-Experts~(MoE)} 
\label{subsec-moe}

%To efficiently improve the model capacity, mixture-of-experts~(MoE)~\cite{shazeer2017outrageously} is proposed to 
We first describe the mixture-of-experts architecture~(MoE)~\cite{shazeer2017outrageously}, which has been used to enhance the model capacity of Transformer based models.
Let $G(x)$ and $E_i(x)$ denote the output vectors of the gating network and the output of the $i$-th expert network for a given input $x$, respectively. 
The output of MoE architecture $y$ can be formally computed as:
\begin{equation}
    y=\sum_{i=1}^{n}G(x) \cdot E_i(x).
    \label{eq-moe-out}
\end{equation}

The $\rm{softmax}$ function is widely adopted as the gating function $G(x)$.
The sparsely-gated MoE architecture, which uses a noisy top-$k$ gating mechanism to reduce the computational cost, has been proposed in~\citet{shazeer2017outrageously}.
It adds tunable Gaussian noise with $H(\cdot)$, and then keeps only the top-$k$ values with $\rm{KeepTopK}(\cdot)$ and sets the rest $-\infty$. This keeps only the top $k$ experts to be evaluated with:
\begin{equation}
    G(x)=\textrm{\rm{softmax}(\rm{KeepTopK}}(H(x),k)).
\end{equation}

Furthermore, Switch Transformer designs a switch routing strategy to simplify this gating function by routing to a \emph{single} expert~\citep{fedus2021switch}. 

\subsection{Tensor and Matrix Product Operators}
\label{sec-MatrixProductOperators}
%\paragraph{Tensor.}
%\ignore{blue}{We refer to  one-dimensional arrays as \emph{vectors},  two-dimensional arrays as \emph{matrices}, and arrays of higher dimensions as \emph{tensors}. Bold uppercase letter~(e.g. $\Matrix{W}$) - matrices, ordinary upper case letters~(e.g. $\Matrix{W}(i,j)$) - matrix elements, calligraphic bold upper case letters~(e.g. $\Tensor{T}$) - for the tensors. In the geometric representation of a tensor, a 3-order tensor can be a representation by a cube.  Tensor and some other related concepts are shown in the appendix.}

%\paragraph{Matrix Product Operators.}

We refer to one-dimensional arrays
as \emph{vectors} (denoted by bold lowercase letters, \eg $\bm{v}$),  two-dimensional arrays as \emph{matrices} (denoted by bold uppercase letters, \eg $\Matrix{W}$), and arrays of higher dimensions as \emph{tensors} (denoted by calligraphic bold uppercase letters, \eg $\Tensor{T}$). 

MPO decomposition~\cite{oseledets2011tensor}~(\emph{a.k.a.} {tensor-train decomposition}) has been a widely used matrix decomposition technique from quantum many-body physics, which  decomposes a matrix~($2$-order tensor) into $m$ local tensors~\cite{pirvu2010matrix}.
Given a matrix $\Matrix{W}_{I\times J}\in \mathbb{R}^{I\times J}$,  the MPO decomposition is given in the following format:
% $\textsc{MPO}~(\Matrix{M})=\prod_{k=1}^n\Tensor{T}_{(k)}[d_{k-1}, i_k, j_k, d_k]$,
\begin{equation}
    \textsc{MPO}~(\Matrix{W})=\prod_{k=1}^m\Tensor{T}_{(k)}[d_{k-1}, i_k, j_k, d_k],
    % \textsc{MPO}~(\Matrix{M})=\prod_{k=1}^{n} \Tensor{T}_{(k)}[d_{k-1},i_k,j_k,d_k],
    \label{eq:mpo}
\end{equation}
where $I = \prod_{k=1}^n i_k$  and $J=\prod_{k=1}^n j_k$,   $\Tensor{T}_{(k)}$ is a $4$-order tensor with size $d_{k-1}\times i_k \times j_k \times d_k$. 
The $d_k$ is dimension of bond linking $\Tensor{T}_{(k)}$ and $\Tensor{T}_{(k+1)}$. 
% For boundary cases, we have $d_0=d_m=1$.
% The $k$-th bond dimension $d_k$ is defined by:
% \begin{equation}
%     d_k = \min\bigg(\prod_{m=1}^k i_m\times j_m, \prod_{m=k+1}^n i_m\times j_m\bigg),
%     \label{eq:d-k}
% \end{equation}
% where $d_0=d_n=1$.
% Then we can obtain the MPO decomposition format of this matrix as follows:
% % $\textsc{MPO}~(\Matrix{M})=\prod_{k=1}^n\Tensor{T}_{(k)}[d_{k-1}, i_k, j_k, d_k]$,
% \begin{equation}
%     \textsc{MPO}~(\Matrix{W})=\prod_{k=1}^n\Tensor{T}_{(k)}[d_{k-1}, i_k, j_k, d_k],
%     % \textsc{MPO}~(\Matrix{M})=\prod_{k=1}^{n} \Tensor{T}_{(k)}[d_{k-1},i_k,j_k,d_k],
%     \label{eq:mpo}
% \end{equation}
% where the $\Tensor{T}_{(k)}$ is a $4$-order tensor with size $d_{k-1}\times i_k \times j_k \times d_k$. 
% 在不对连接键dk做截断的情况下，MPO(W)通过张量缩并后可以得到原矩阵W。

\begin{figure}[t]
    \centering
    \includegraphics[width=0.46\textwidth]{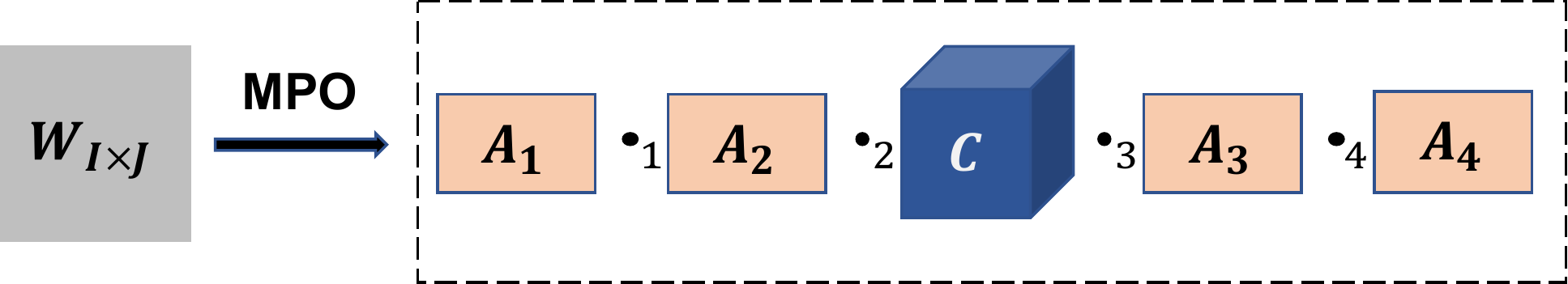}
    \caption{MPO decomposition for matrix $\Matrix{W}_{I\times J}$ with five local tensors.
    Auxiliary tensors~($\{\Tensor{A}_{i}\}_{i=1}^{4}$) and central tensor~($\Tensor{C}$) are marked in orange and blue, respectively. }
    \label{fig:mpo_d}
\end{figure}

According to \citet{gao2020compressing}, the original matrix $\Matrix{W}$ can be exactly reconstructed by tensor contraction of MPO$(\Matrix{W})$ without truncation of the connection bond $\{d_k\}_{k=1}^m$.
Figure~\ref{fig:mpo_d} presents the illustration of the MPO decomposition procedure for a matrix~($m=5$). More detailed analysis on different factorization ways (\ie $m=3,5,7,9$) will be given in Section~\ref{sec-detailed_analysis}.
%and we use $n=5$ in our experiments for convenience.
% We present a detailed algorithm for MPO decomposition in Algorithm~\ref{alg:mpo-decomposition}.
% $``\bullet_{k}"$ is denoted as the tensor-tensor product on the $k$-th order~\cite{kolda2009tensor} and $k\in\{1,\dots,d\}$.
% The bond dimension $d_k$ is defined by:
% \begin{equation}
%     d_k = \min\bigg(\prod_{m=1}^k i_m\times j_m, \prod_{m=k+1}^n i_m\times j_m\bigg).
%     \label{eq:d-k}
% \end{equation}
% From Eq.~\eqref{eq:d-k}, we can see that is going to be large in the middle and small on both sides. We present a detailed algorithm for MPO decomposition in Algorithm~\ref{alg:mpo-decomposition}.
After MPO decomposition, the central tensor~(the tensor right in the middle) with most of the parameters can encode the core information of the original matrix, while the auxiliary tensors~(the rest of these tensors) with only a small proportion of parameters play the role of complementing the central tensor.
% It is usually preferred to use odd local tensors with MPO decomposition.

% As we will discuss in the experimental~\ref{}, the capability of MPO does not change depending on the $n$.

% 给定任意矩阵都可以做MPO分解，并且其分解形式不唯一。

% In this case, we refer to the tensor right in the middle as \emph{central tensor}, and the rest as \emph{auxiliary tensor} following~\citep{DBLP:conf/acl/LiuGZXLW20}.
% A merit of MPO decomposition is that the majority of information of the original matrix is aggregated in the central tensor~(with most of the parameters), and tuning auxiliary tensors~(with only a small proportion of parameters) while fixing central tensor can yield large change in original matrix~\citep{DBLP:conf/acl/LiuGZXLW20}. 

\section{Approach}
\label{sec-approach}

To reduce the information redundancy across different experts, we design an MPO-based MoE architecture for increasing the model capacity in a parameter-efficient way. 
We firstly describe the MPO-based MoE architecture and then introduce an improved optimization algorithm for learning the parameters in this architecture. 
% MoE can increase the model capacity of Transformer-based models, by incorporating additional FFN layers~(mainly in the form of a matrix) in Transformer networks as experts~\cite{shazeer2017outrageously}. 

\subsection{MPO-based Mixture-of-Experts}
\label{sec-MPOE}
% 由于专家层内的参数高耦合，以往的MoE模型的需要训练专家层中所有的参数，并且通过不同的路由函数进行专家的筛选。这导致了模型扩展的参数不高效。
Previous MoE architecture~\cite{jacobs1991adaptive,shazeer2017outrageously} usually treats different experts as individual components, requiring a compete copy of network parameters for each expert. Although it has been found~\cite{fedus2021switch,kim2021scalable}  that there exists redundant information among different experts in the MoE architecture, it is not easy to identify the shareable parameters from the highly coupling network. %A major limitation of these studies is that they seldom consider structural decomposition of parameter matrices 
%Due to the high coupling of parameters within the expert layers, previous MoE architecture~\cite{jacobs1991adaptive,shazeer2017outrageously} treats different experts as individual modules and keeps a considerable amount of parameters in each expert. 
%Such a way is likely to lead to information redundancy across experts since the experts are learned with correlated data samples in training or pre-training datasets.
% ~指出在不同专家之间具有荣有冗余性信息，因此在使用稀疏模型部署的时候，采用蒸馏或者裁减掉冗余专家的方法来改善这个问题。
%~\citet{fedus2021switch,kim2021scalable} pointed out that there is redundant information among different experts, so distillation or pruning of redundant experts is used to improve this in deployment using sparse models.
%We also found cross-expert redundancy features in the MoE framework in our experiments by inputting the same samples into different experts after fine-tuning and calculating the maximum mean discrepancy~\citep{2007mmd} between the output of the features by different experts. we found that these features by different experts conform to the same distribution, which illustrates the existence of information redundancy between different experts.
% ~(see Appendix.~\ref{appendix-mmd-in-experts} for details).

Considering this issue, our solution is inspired by an important merit of MPO decomposition: it can reorganize and aggregate the core information in central tensors~\cite{gao2020compressing} as aforementioned. Based on this property, the core idea of our approach is to share the central tensors for all the expert layers and enable specificity via expert-specific auxiliary tensors. 
%Next, we conduct an empirical analysis to examine the information redundancy across experts and then present the proposed MPO-based MoE architecture. 

\begin{figure*}[ht]
\centering
\subfigure[MPO-based mixture-of-experts architecture.]{
\begin{minipage}[b]{\columnwidth}
\centering
\includegraphics[width=0.9\columnwidth]{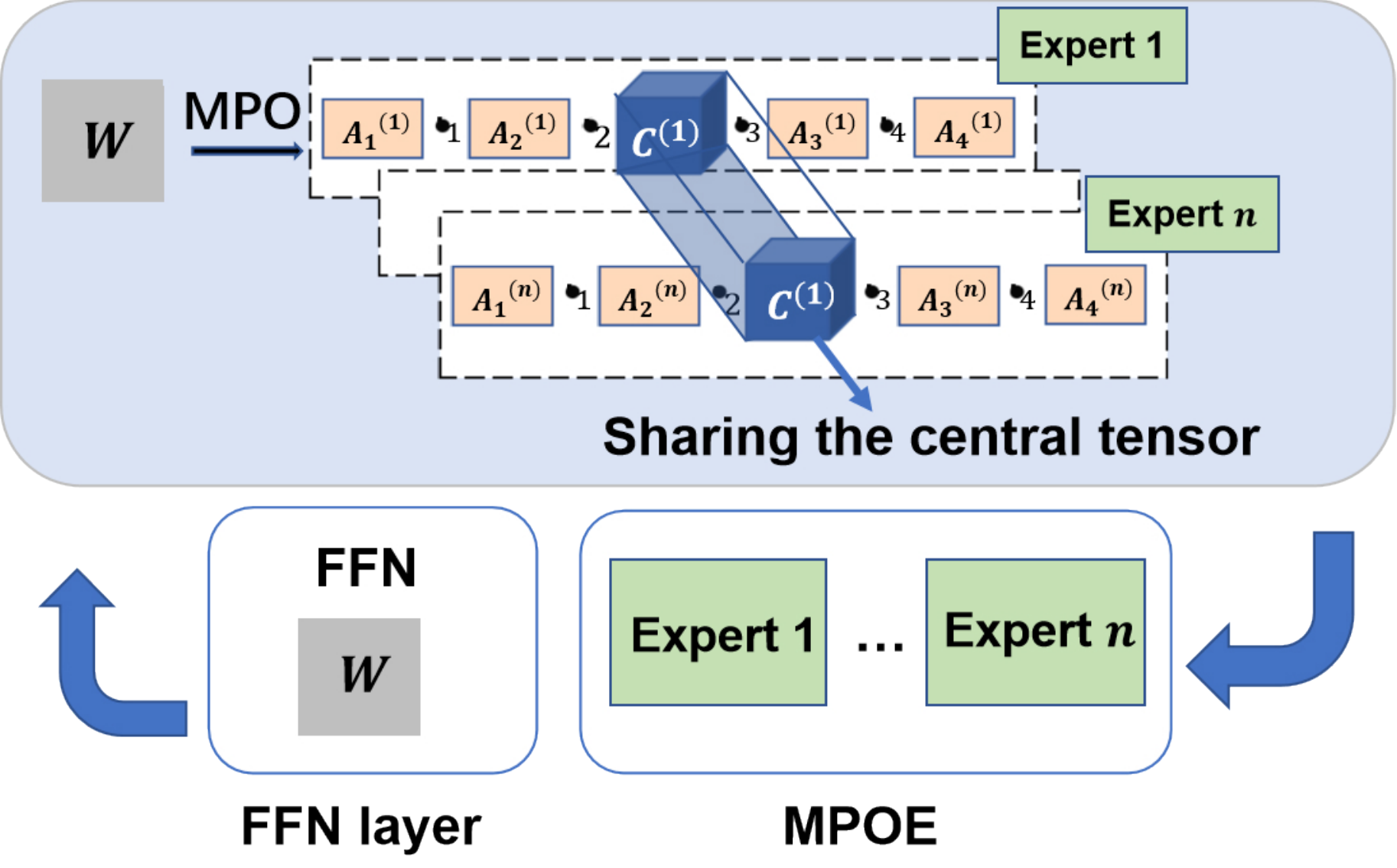} 
\end{minipage}%
}%
% \hspace{-0.5cm}
\subfigure[Gradient mask strategy.]{
\begin{minipage}[b]{\columnwidth}
\centering
\includegraphics[width=\columnwidth]{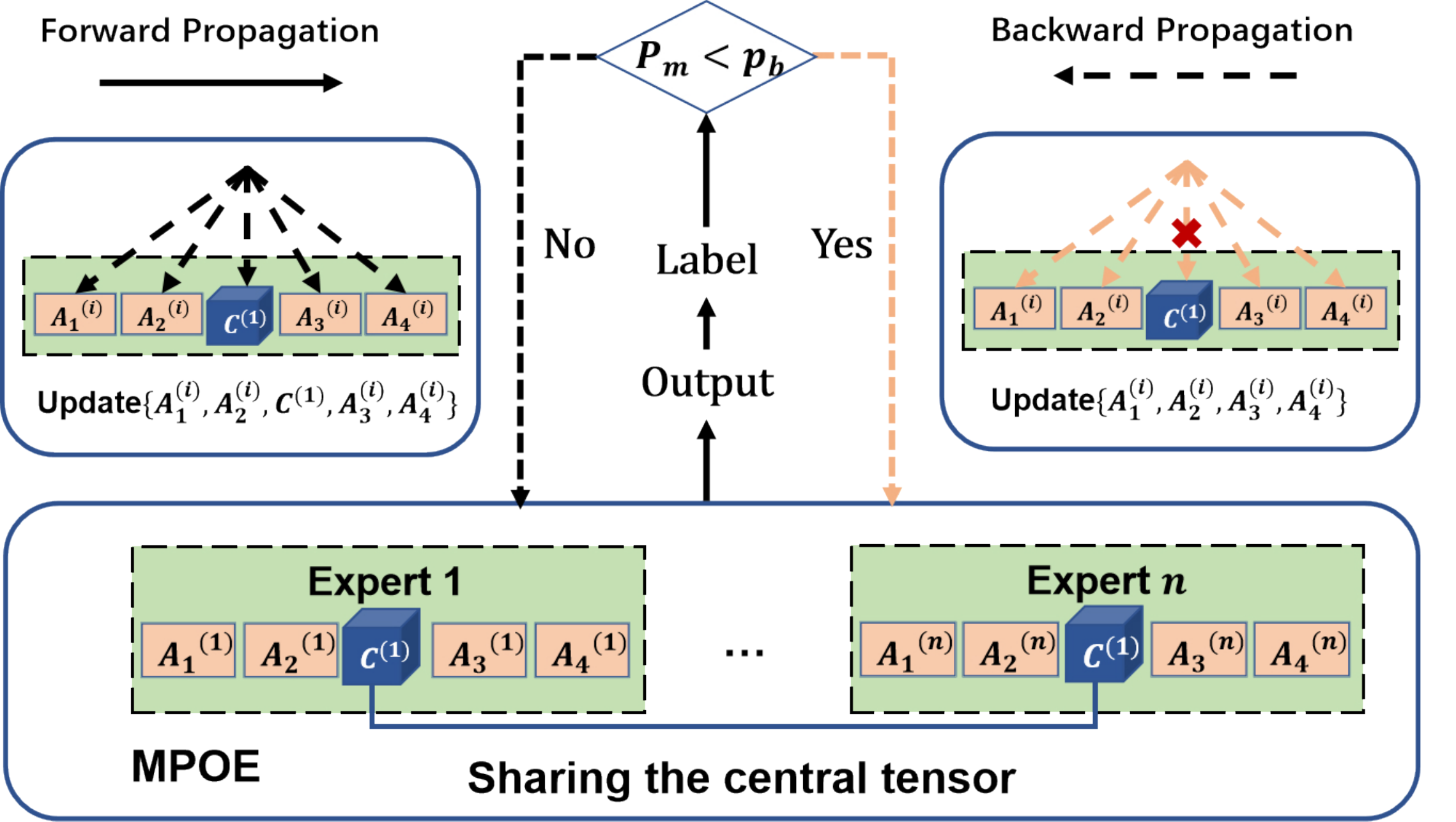} 
\end{minipage}%
}%
\caption{Illustration the proposed MPOE architecture and gradient mask strategy. 
    We decompose the weight matrix of each expert in the MoE architecture into five local tensors using MPO, containing four auxiliary tensors and one central tensor, which are marked in orange and blue, respectively.    In our approach, the central tensor of the $n$ experts is shared in the MPOE architecture.
   During optimization, each backward propagation process updates a set of auxiliary tensors while updating the central tensor with a probability of $p_b$~(the mask probability of the central tensor), which can effectively avoid the unbalanced optimization of the central tensor.
    }
\label{fig:mpoe_layer}
\end{figure*}
% \begin{figure*}
%     \centering
%     \includegraphics[width=1.0\textwidth]{image/model_main_5.pdf}
%     \caption{\textcolor{blue}{use a/b subfigures}
%     (a) 
%     Illustration of the proposed MPOE architecture. 
%     We decompose the weight matrix of each expert in the MoE architecture into five local tensors using MPO, containing four auxiliary tensors and one central tensor, which are marked in orange and blue, respectively.    Technically, the central tensor of the $n$ experts is shared in the MPOE architecture.
%     ~(b)
%     Illustration of gradient mask strategy. $p_d$ is the mask probability of the central tensor.
%     \textcolor{blue}{rewriting the whole caption}
%     }
%     \label{fig:mpoe_layer}
% \end{figure*}

\paratitle{Parameter-Efficient MoE Architecture}.
\label{subsubsec-lightnetwork}
The Transformer network consists of two major neural components, namely FFN and multi-head attention. 
% As depicted in~\cite{shazeer2017outrageously,fedus2021switch}, we extend the FFN layers as experts. 
Following previous work on MoE-based PLMs~\citep{shazeer2017outrageously,fedus2021switch}, we consider FFN layers as experts to be extended, while our approach is generally applicable to various matrix-based model components. 
A straightforward method to reducing information redundancy  is to share a proportion of parameters across experts. However, in Transformer-based networks, the experts~(\ie FFN here) are mainly composed of large dense matrices, which are difficult for sharing partial parameters from these matrices. As our solution, we consider parameter sharing through the MPO decomposition, so that the derived central tensors can be flexibly shared across matrices.

\paratitle{Lightweight MoE Design}. Specifically, we simplify the discussion by assuming that an expert corresponds to one parameter matrix at each layer, and it is similar for the multi-matrix cases.
We consider a MoE architecture of $n$ experts each with $L$ layers, so that there are $L \times n$ matrices in total, denoted by $\{\Matrix{W}^{(l,i)}\}_{l=1,i=1}^{L,n}$. As discussed in Section~\ref{sec-MatrixProductOperators}, a matrix can be decomposed into $m$ tensors, consisting of one central tensor and $m-1$ auxiliary tensors.
In this work, we consider five decomposed tensors, \ie $m=5$. At the $l$-th layer, the decomposition results can be denoted by $\{\Tensor{C}^{(l,i)}, \Tensor{A}_1^{(l,i)},\Tensor{A}_2^{(l,i)},\Tensor{A}_3^{(l,i)},\Tensor{A}_4^{(l,i)}\}_{i=1}^{n}$, where $\Tensor{C}^{(l,i)}$ and $\Tensor{A}_{\cdot}^{(l,i)}$ are the central and auxiliary tensors of the $i$-th parameter matrix, respectively, at the $l$-th layer.
To develop the MPO-based MoE architecture, the core idea is to share the central tensors as global parameters and keep expert-specific auxiliary tensors as local parameters, \ie 
$\Tensor{C}^{(l,1)}=\Tensor{C}^{(l,2)}\cdots =\Tensor{C}^{(l,n)}$ ($\forall~l=1\cdots L$), and we denote the global central tensor at the $l$-th layer by $\Tensor{C}^{(l)}$. In this way, we can only keep $L$ central censors for a $L$-layer MoE architecture. 
For MPO, the decomposition process is transparent to external modules, so that we can reuse the previous routing mechanism~(Section~\ref{subsec-moe}) by distributing data samples to different experts. A slight difference is that we only need to consider the routing to local tensors for each matrix since the global tensor is shared across experts. 
We call such an MPO-based MoE architecture as \rm{\bf{MPOE}}.

%With the MPO decomposition on a given matrix, the derived central tensor covers the major proportion of parameters so that our approach can largely reduce the redundant parameters. 
%Although it is appealing to directly share the central tensor, a key question about the effectiveness of this approach is whether the current architecture enables sufficient flexibility and diversity for each expert. To answer this question, we refer to an important property of MPO decomposition from many-body physics~\cite{}: there is close entanglement between central and auxiliary tensors, and even small parameter variations from auxiliary tensors can significantly affect central tensors.

\paratitle{Discussion}. 
% After MPO decomposition, \textcolor{blue}{the number of tensor parameters is positively correlated with the capacity of information,} so that the information capacity of the central tensor is the largest. 
Since the central tensor contains most of the information from original parameter matrices~\cite{gao2020compressing}, a key question is whether the current architecture enables sufficient flexibility and specificity for each expert. To answer this question, we refer to an important property of MPO decomposition from quantum many-body physics~\cite{pirvu2010matrix}: 
it is guaranteed in principle, that any change on one tensor will be propagated to the entire local tensor set. 
In other words, only tuning the auxiliary tensors~(keeping the central tensor fixed) can lead to the same effect as tuning the whole matrix. 
Since the parameters of the central tensor are shared, our approach can significantly reduce the number of actual parameters given the MoE architecture with the same number of experts. 
Assuming the original model consisting of $n$ experts with $T$ parameters each, we have a total number of $n\cdot T$ parameters. 
Specifically, let $\gamma$ denote the parameter ratio of the auxiliary tensor to the central tensor for expert networks. Given the total number $T$ for an expert network, the central and auxiliary tensors correspond to the parameters numbers of $\frac{\gamma}{\gamma+1} T$ and $\frac{1}{\gamma+1}T$, respectively. Since our MPOE approach shares the central tensor, the final number of parameters will be $ \frac{\gamma}{\gamma+1} T + \frac{n}{\gamma+1} T $. Thus, our MPOE approaches corresponds to a ratio of $\frac{n+\gamma}{n(\gamma+1)}$ of the original parameter scale. In our experiments, the ratio $\gamma$ is about 12, and $\frac{n+\gamma}{n(\gamma+1)}$ approximately equals to $0.19$ when  $n=8$. Such a ratio will be further decreased when we have more experts. It can be seen that our MPOE approach is able to effectively reduce the parameter scale.

\subsection{Alleviate Unbalanced Optimization}
% Fast Training with Expert Selection.
\label{subsec-fast training}
As the experts share the central tensor in the MPOE approach, the corresponding parameters of the central tensor will be updated more frequently than those in the auxiliary tensors during fine-tuning.
It tends to lead to the unbalanced optimization issue as reported by~\citet{xu2021child}, due to deviation from the pre-trained weights.
As a result, it is crucial to develop a more stable optimization technique that is suited to the MPOE architecture.

Inspired by the solution of gradient dropout strategy~\cite{tseng2020regularizing,xu2021child}, we propose to mask the gradients for the central tensor to improve model optimization for the MPO-based MoE architecture.
At each iteration, we take a certain probability $p_b$ to discard the update in the central tensor. This can effectively alleviate the unbalanced optimization which is caused by the frequent updates of the central tensor.
Specifically, we generate a binary mask $b$ drawn from Bernoulli distribution with a mask probability $p_b$, which can be calculated by $b \sim {\rm{Bernoulli}}(p_b)$. 
% pb越大，代表中心张量更新的次数就越频繁，当pb等于1的时候，表示对于每个样本输入都更新中心张量的参数。
We denote the $\Delta \Tensor{C}$ as the update of the central tensor at each iteration:
\begin{equation}
    \Delta \Tensor{C} = \eta \frac{\partial \mathcal{L}(\Tensor{C})}{\partial \Tensor{C}} \odot (1-b).
\end{equation}
The larger $p_b$ is, the less frequently the central tensor is updated. 
In particular, when $p_b$ is equal to 1, it means that the parameters of the central tensor are frozen for each input of the data. The computational cost of central tensor update can be also reduced with this trick.
% 同时，辅助张量的梯度不做掩码操作，进行标准的梯度更新。我们在算法1中对整个更新过程进行了详细的描述。

Note that the gradient mask trick is only applied to central tensors. For auxiliary tensors, we perform the standard gradient update for learning the parameters. 
Compared with two alternative ways to implement the gradient mask technique, \ie mask pre-activation or post-activation in FFN layers, we find that such a sampling-based masking strategy can effectively improve the model performance in our experiments. 
% There are alternative ways to implement the gradient mask technique, \eg  mask pre-activation or post-activation in FFN layers. 
% In our experiments, we find that such a sampling-based masking strategy can effectively improve the model performance. 
% We will discuss the comparison of different gradient mask methods in detail in the Appendix. 
%The default decision for our approach is to mask all the central tensors in MPOE. 
%All our experiments use this default decision unless otherwise specified. We compare this design decision against other strategies in the appendix.

\subsection{The Overall Algorithm Procedure}

Our approach can be generally applied to various MoE-based models for increasing the model capacity.
In this work, we adopt the MoE-extended PLMs~\cite{radford2019language} for study.

Algorithm~\ref{alg-overall-process} presents a complete procedure for the proposed update procedure, which can be briefly summarized as follows. 
First, we obtain the PLM and perform MPO decomposition for each weight matrix of the FFN layers in the Transformer.
For each weight matrix, we decompose it into one central tensor~$\Tensor{C}$ and a list of auxiliary tensors~$\Tensor{A}$.
In the original MoE architecture, we will have $n$ sets of such decomposed parameters.
Next, the key point lies in that we share the central tensor $\Tensor{C}$ in the decomposition process but keep expert-specific auxiliary tensors. 
% Next, we share the central tensor across  $n$ experts, $\ie$  one global central tensor, and expert-specific auxiliary tensors.
In this way, each expert is composed of a set of auxiliary tensors and a shared central tensor.
To recover the original FFN matrix in some specific expert, we can simply multiply the shared central tensor by expert-specific auxiliary tensors. 
% \textcolor{blue}{While, during the training process, we will always update the shared central tensor for each sample, but only modify the corresponding set of auxiliary tensors~(depending on the gating function). }
Then, we apply the gradient mask strategy to update the parameters in these experts, $\ie$  masking the gradient of the central tensor.

Since the parameters of the central tensor are two orders of magnitude larger than the parameters of the auxiliary tensors~\citep{DBLP:conf/acl/LiuGZXLW20}, the cost of MoE-based networks will be largely reduced by sharing the central tensor. 

% \begin{algorithm}[ht]
%     \caption{The proposed update procedure.}
%     \begin{algorithmic}[1] %每行显示行号
%         \Require  $\Matrix{W}$: initial pre-trained weight; 
%         $\mathcal{L}(\Tensor{C})$: stochastic objective function with central tensor; $\mathcal{L}(\Tensor{A})$: stochastic objective function with auxiliary tensor; 
%         $\eta$: learning rate; $p_d$: mask probability; $n$: number of experts;
%         \State \textbf{initialize}: time step $t\gets 0$ 
%         \State $\{\{\Tensor{A}_j\}_{j=1}^4,\Tensor{C}\} \gets {\rm{MPO}}(\Matrix{W}) $ 
%         \For {$i = 1 \to n$}
%             \State Initialize $\{\Tensor{A}_j\}_{j=1}^4$;
%         \EndFor
%         % \State Initialize $\{\Tensor{A}_1,\Tensor{A}_2,\Tensor{A}_3,\Tensor{A}_4\}_{i=1}^N$
%         \State Sharing the central tensor $\Tensor{C}$
%         \State $\Tensor{C}^{t=0}\gets \Tensor{C}, \quad \Tensor{A}^{t=0} \gets \Tensor{A}$
%         \While {not converged} 
%             \State $t \gets t+1$
%             \State $g_{\Tensor{C}}^{t} \gets \frac{\partial\mathcal{L}(\Tensor{C}^t)}{\partial(\Tensor{C}^t)}$,\quad $g_{\Tensor{A}}^{t} \gets \frac{\partial\mathcal{L}(\Tensor{A}^t)}{\partial(\Tensor{A}^t)}$
%             % \State $P_m \gets {\rm{GenerateMask}}(p_b)$
%             \State $b \gets {\rm{GenerateMask}}(p_d)$
%             \State $\Tensor{C}^t \gets \Tensor{C}^{t-1} - \eta \cdot g_{\Tensor{C}}^t\odot (1-b)$
%             \State $\Tensor{A}^t \gets \Tensor{A}^{t-1} - \eta \cdot g_{\Tensor{A}}^t$
%         \EndWhile
%     \end{algorithmic}
% \label{alg-overall-process}
% \end{algorithm}
\begin{algorithm}[ht]
    \caption{The proposed updating procedure.}
    \begin{algorithmic}[1] %每行显示行号
        % \Require  $\Matrix{W}$: initial pre-trained weight 
        \Require $\{\{\Tensor{A}_j\}_{j=1}^4,\Tensor{C}\}$: Initialize experts
        \Require $\alpha$: learning rate
        \Require $p_b$: mask probability
        \Require time step $t\gets 0$~(Initialize timestep)
        \While {not converged} 
            \State $t \gets t+1$
            \State $g_{\Tensor{C}}^{t} \gets \frac{\partial\mathcal{L}(\Tensor{C}^t)}{\partial(\Tensor{C}^t)}$,\quad $g_{\Tensor{A}}^{t} \gets \frac{\partial\mathcal{L}(\Tensor{A}^t)}{\partial(\Tensor{A}^t)}$
            \Statex (Get gradients at timestep $t$)
            \State $b \gets {\rm{GenerateMask}}(p_b)$
            \Statex (Compute gradient mask)
            \State $\Tensor{C}^t \gets \Tensor{C}^{t-1} - \alpha \cdot g_{\Tensor{C}}^t\odot (1-b)$
            \Statex (Update central tensors)
            \State $\Tensor{A}^t \gets \Tensor{A}^{t-1} - \alpha \cdot g_{\Tensor{A}}^t$
            \Statex (Update the routed auxiliary tensors)
        \EndWhile
        \State \Return $\{\{\Tensor{A}^t_j\}_{j=1}^4,\Tensor{C}^t\}$~(Resulting parameters)
    \end{algorithmic}
\label{alg-overall-process}
\end{algorithm}

\subsection{Discussion}

For the parameter inefficiency issue of MoE-based networks,  
existing studies mainly focus on alleviating the unbalanced load of experts, which have proposed different 
routing methods to balance the routing probabilities of different experts, such as BASELayer~\cite{lewis2021base}, HASHLayer~\cite{roller2021hash}, GShard~\cite{dmitry2021gshard} and Switch Transformers~\cite{fedus2021switch}. 
As a comparison, we aim to reduce information redundancy by sharing common parameters among experts. Actually, the MPOE approach can be further enhanced with existing improved routing methods.   

Specifically, Deepspeed-MoE proposed to use pyramid residual MoE architecture to reduce the parameters of the MoE architecture~\cite{rajbhandari2022deepspeed}, while our work takes a different perspective to improve the original MoE architecture by sharing parameters among different experts.
% However, our work and Deepspeed-MoE take very different approaches to improving the original MoE architecture: Deepspeed-MoE~(pyramid residual MoE architecture) and our work~(MoE-based parameter sharing).

% \textcolor{blue}{For applications using MoE to enhance PLM, recent work focus on adapting MoE to current PLM or distributed training system to accelerate model training or fine-tuning, such as FastMoE~\cite{he2021fastmoe} and MoEfication~\cite{zhang2021moefication}. 
% The proposed MPOE approach can also be used on these MoE-enhanced PLMs.
% }

\section{Experiments}

\begin{table*}[t]

\small
\centering
\begin{tabular}{lrrrrrrrr|rr}
\bottomrule
\rowcolor{gray!10}\multicolumn{11}{l}{\it \textbf{NLU with T5}}                                                                                               \\

\toprule[1pt]

Experiments   & MNLI	  & QNLI     & SST-2  	  &RTE       &QQP      &CoLA         &MRPC	       &STS-B                    &Avg.            & {\#To~(M)} \\ \midrule
T5-Large     & \textbf{89.23}  & 94.03    & 96.20       & 83.94    & \textbf{91.54}   & 55.10        & 90.15       & 91.90               & 86.51          &         737       \\ 
+MoEfication$\blacklozenge$  & 87.50   & 93.20     & 95.40       & 86.40     & 90.20    & 55.50        & 87.50        & 90.60               & 85.79            &       737        \\ 
+MoEfication$_{++}$ $\blacklozenge$ & 88.70   & 93.60     & 96.20       & 87.50     & 91.30    & 59.40        & 89.30        & 91.00               & 87.13           &        737      \\ 
+Switch$\clubsuit$& / & /     & /       & /    & /  & /      & /        & /                                                   & 88.50                 & 26000        \\ 
+MPOE        & 87.16   & \textbf{94.12}    & \textbf{96.80}        & \textbf{88.60}     & 90.63   & \textbf{67.63}     & \textbf{93.65}       & \textbf{91.97}             & \textbf{88.82}              & 956          \\  \midrule
T5-Base      & 87.78  & 93.82    & 94.72       & 71.74    & 91.11   & 53.49      & 89.16       & 91.16              & 84.12               &  223      \\ 
+Switch$\clubsuit$& / & /     & /       & /    & /  & /      & /        & /                                                   & 86.70                &  3800        \\ 
+Switch$\spadesuit$      & 87.73  & 93.85    & \textbf{94.87}       & \textbf{77.53}   & 91.59   & 59.90      & 91.64       & 91.16             & 86.03              &  1015         \\
+MoE$\bigstar$         & 86.98  & 92.82    & 94.60        & 69.56    & 90.02   & 64.56      & 87.68       & 90.89             & 84.64               &     1015     \\
+MPOE        & 87.60   & 93.30     & 94.81       & 77.13    & 90.81   & 65.53      & \textbf{93.14}       & 91.17             & 86.69               &   294       \\
+MPOE$_{++}$       & \textbf{87.78}  & \textbf{93.93}    & 94.83       & 77.42    & \textbf{91.61}   & \textbf{65.90}      & 91.14       & \textbf{91.65}         & \textbf{86.78}               &   365       \\ \bottomrule
\end{tabular}

\small
\centering
\begin{tabular}{lrrrrrrr|r}
\rowcolor{gray!10}\multicolumn{9}{l}{\it \textbf{NLG with GPT-2}}                                                                                               \\
\toprule[1pt]
  \multicolumn{1}{c}{\multirow{2}{*}{Experiments}} &
  \multicolumn{3}{c}{WikiText-2} &
  \multicolumn{2}{c}{EMNLP News} &
  \multicolumn{2}{c}{IMDB} &
  \multicolumn{1}{c}{\multirow{2}{*}{\#To~(M)}} \\
                     & PPL~($\downarrow$)            & BLEU-2          & BLEU-4          & BLEU-2          & Self-BLEU-2    & BLEU-2      & Self-BLEU-2     & \multicolumn{1}{c}{} \\ \midrule
GPT-2    & 21.27          & 28.69           & 9.46            & 62.61           & 74.67          & 73.12         & 83.85         & 124                           \\ 
+MoE$\bigstar$                  & 21.86          & 28.27           & 9.14            & 65.27           & 79.79          & 74.46         & 90.01         & 578                             \\
+Switch$\spadesuit$               & 21.25          & 28.71           & 9.44            & 64.62           & 81.11          & 75.35         & 91.82         & 578                            \\
+MPOE                 & \textbf{20.72}          & 28.78           & 9.51            & 66.99           & 83.10           & 76.30& 92.72& 157                \\ 
+MPOE$_{++}$               & 20.73 & \textbf{28.82}  & \textbf{9.57}   & \textbf{68.49}  & \textbf{83.11}          & \textbf{76.82}          & \textbf{93.08}         & 171         
\\\bottomrule
\end{tabular}
\caption{Performance comparison of different models on  NLU and NLG tasks (in percent).  ``\#To~(M)'' denote the number~(in millions) of total parameters.
We set the number of experts $n=8$ in these models, MPOE.
Furthermore, we use $n=16$ for a more powerful version of our approach, denoted by MPOE$_{++}$.
We report the average test performance of three runs, and the best results are highlighted in bold.
$\blacklozenge$: Experimental results by~\citet{zhang2021moefication}
$\clubsuit$: Experimental results by~\citet{fedus2021switch}
$\spadesuit$: Our re-implementation by~\citet{fedus2021switch}.
$\bigstar$:Apply method by~\citet{shazeer2017outrageously}.
}
\label{tab:main_results}
\end{table*}

In this section, we first set up the experiments and then report the results and analysis.
Then, we conduct a detailed analysis under different experimental settings.
Here, 
%Note that MPOE can be applied generally to all the existing PLMs, and 
we use T5~\cite{multitask2020raffel} and GPT-2~\cite{radford2019language} models as the base model in our experiments.
\subsection{Experimental Setup}
\label{sec-experimental-setup}
\paragraph{Datasets.} 
To evaluate the effectiveness of the proposed MPOE as an efficient strategy to improve the model capacity of PLMs, we follow the setting of T5 and GPT-2 to perform experiments on Natural Language Understanding~(NLU) and Natural Language Generation~(NLG) tasks.
Specifically, we evaluate the NLG tasks in GLUE benchmark~\cite{wang2018glue}, the language modeling task with WikiText-2~\cite{wikitext}, the text generation task with IMDB~\cite{imdb} and EMNLP2017 WMT News~\cite{EMNLPnews}.
Furthermore, we follow the setup of~\citet{multitask2020raffel} on the GLUE benchmark for a direct comparison with the T5 model. 

GLUE benchmark covers multiple datasets (MRPC, QQP, SST-2, MNLI, RTE, QNLI, CoLA)\footnote{Following~\citet{multitask2020raffel}, as a common practice, due to the adversarial nature of WNLI with respect to the training set, we do not experiment with WNLI}. The original test sets are not publicly available, and following~\citet{fewsample2021zhang}, for datasets fewer than 10$K$ samples~(RTE, MRPC, STS-B, CoLA), we divide the original validation set in half, using one half for validation and the others for the test.
% For metrics used in the GLUE benchmark, we follow~\cite{rabeeh2021hyperformer} and use Matthew's correlation for CoLA, Pearson for STS-B, and accuracy for the other tasks.
\paragraph{Evaluation Metrics.}
We use perplexity~(PPL)~\cite{DBLP:journals/coling/BrownPPLM92} to measure how well the probability model predicts a sample compared with the ground-truth. To evaluate the ratios of the overlapping $n$-grams between generated and real samples, we use BLEU-$n$ score~\cite{DBLP:conf/acl/PapineniRWZ02}. We also take into account the Self-BLEU-$n$ score~\cite{DBLP:conf/sigir/ZhuLZGZWY18} to evaluate the diversity of generated samples especially.
% py
For metrics used in the GLUE benchmark, we follow~\citet{rabeeh2021hyperformer} and use Matthew's correlation for CoLA, Pearson for STS-B, and accuracy for the other tasks.

\paragraph{Comparison methods.}
% We use the GPT-2 as the default architecture for both MoE and MPOE. 
We adopt the T5 and GPT-2 as the base architectures for both MoE and MPOE. 
Following~\citet{shazeer2017outrageously}, we extend the FFN components with the MoE architecture containing $n$ experts in each Transformer block of the T5 and GPT-2 model. We refer to this method as ``+MoE''. 
The Switch Transformers~\cite{fedus2021switch} use a simplified strategy that routes to only a single expert instead of top-$2$ routing in MoE.
We refer to this method as ``+Switch''. 
To ensure a fair comparison, we maintain the same number~($n=8$) of experts for baselines and MPOE. 
We also implement an enhanced  version of MPOE with $n=16$ experts, which is referred to as ``+MPOE$_{++}$''.  
Based on the released \emph{gpt2} model\footnote{https://huggingface.co/gpt2}, \emph{t5-base} model\footnote{https://huggingface.co/t5-base} and  \emph{t5-large} model\footnote{https://huggingface.co/t5-large} provided by Huggingface, 
we first initialize the experts, then fine-tune the models on the downstream tasks.
% \textcolor{blue}{For multi-task fine-tuning experiments, we follow the setting in~\citet{rabeeh2021hyperformer}. (what do you follow?)}
% For multi-task fine-tuning, we follow the setting in~\citet{rabeeh2021hyperformer} and fine-tune all parameters of the model on all tasks.
For the T5 model, we follow the setting in~\citet{rabeeh2021hyperformer} and fine-tune all parameters of the model on all tasks. 
% We refer to the single-task training and multi-task training as ``ST'' and ``MT'', respectively.
For different downstream tasks, we run a hyperparameter sweep and select the best configuration according to the accuracy results on the validation set. The hyperparameters that we tune include the epochs, batch size and learning rates.

% All the experimental codes will be released after the review period.
% We will release all the experimental codes after the review period.
\subsection{Mains Results}
\label{sec-exprimental results}
%s\paragraph{Comparison with MoE Variants.}
In our main experiments, we adopt T5~\cite{multitask2020raffel}, GPT-2~\cite{radford2019language}, Switch Transformers~\cite{fedus2021switch} and MoEfication~\cite{zhang2021moefication} as baselines, and report the comparison results  of both NLU and NLG tasks in Table~\ref{tab:main_results}. 

% % 总体来说，相较于这些MoE变种，对于NLU任务，T5+MPOE++在GLUE的基准下达到了最好的效果；对于NLG任务，GPT-2+MPOE achieves competitive performance with reduced model sizes on these three datasets
% Overall, compared to these MoE variants, for the NLU task, T5+MPOE$_{++}$ achieves the best results on the GLUE benchmark; for the NLG task, GPT-2+MPOE achieves competitive performance with reduced model sizes on these three datasets.
% \textcolor{blue}{(explain why)}
% % --- lpy
% By zooming into specific tasks, our proposed approach (``+MPOE'') outperforms the best baseline method (\ie ``+Switch'')~(88.82 vs. 88.50 for T5-Large and 86.78 vs. 86.70 for T5-Base) with up to 27.2x reduction in total parameters~(956M vs. 26,000M for T5-Large) in GLUE benchmark.
% % --- lpy
% % with {96.3\%} total parameter size reducing~(26,000M vs. 956M).
% % with \textcolor{blue}{{96.3\%} total parameter size reducing~(26,000M vs. 956M).}
% GPT-2+MPOE achieves BLEU-2 gains~(1.72 for GPT-2+MoE and 2.37 for GPT-2+Switch) with 3.7x reduction in total parameters~(578M vs. 157M) on EMNLP News dataset.
% ----- 0906 lpy
Overall, compared to these MoE variants, our proposed MPOE approach achieves performance improvement while being more parameter-efficient. 
For the NLU task, our proposed approach (``+MPOE'') outperforms the best baseline method, \ie ``+Switch''~(88.82 vs. 88.50 for T5-Large) with up to 27.2x reduction in total parameters in the GLUE benchmark. 
% By zooming into low-resource datasets such as CoLA and MRPC, our approach yields more significant improvements, which indicates positive transfer effects of sharing information across experts.
By zooming into low-resource datasets such as CoLA and MRPC, our approach yields more significant improvements. This suggests that sharing parameters across experts reinforces the positive transfer effects\footnote{Here, the positive transfer effects can be referred to~\citet{rabeeh2021hyperformer}, which means that the transferred knowledge can lead to improved performance for unseen in-domain tasks.} of information from other datasets toward the learning of low-resource datasets.
For the NLG task, GPT-2+MPOE achieves gains in BLEU-2 score~(1.72 for GPT-2+MoE and 2.37 for GPT-2+Switch) with 3.7x reduction in total parameters on the EMNLP News dataset. This indicates that GPT-2 also benefits from sharing central tensors.

Moreover, T5+MPOE$_{++}$ and GPT-2+MPOE$_{++}$ perform better when we add more auxiliary tensors as additional experts.
% when we add more experts by including the auxiliary tensors,
This demonstrates the necessity of improving model capacity~\cite{shazeer2017outrageously}, as more parameters of experts tend to result in an improved model capacity.

\subsection{Evaluation on Multi-task Learning}
% 在多任务学习中，我们为了与（）可比，因此我们采用了T5模型进行分析。
To demonstrate the efficiency of MPOE in multi-task learning, we adopt the T5-Base model for analysis to be comparable with Hyperformer~\cite{rabeeh2021hyperformer}.
We conduct experiments on the multi-task GLUE benchmark.
The detailed metrics can be found in Section~\ref{sec-experimental-setup}.
% Note that compared to Hyperformer, MPOE does not require \textcolor{blue}{any additional structure, it is more applicable to the basic model.  
Note that compared to Hyperformer, MPOE approach does not incorporate additional neural network components, thus it is more flexible to be used with the PLMs.

% Table~\ref{tab:multi-task} shows the results on GLUE benchmark for single-task and multi-task training.}
Table~\ref{tab:multi-task} shows the results on GLUE benchmark for T5-base~\cite{multitask2020raffel}, Hyperformer~\cite{rabeeh2021hyperformer} and MPOE.
As we can see, the performance of the MPOE approach is consistently better than the Hypernetwork in all cases, while the MPOE is more parameter-efficient~(258M vs. 343M in total parameters).
% Compared to single-task fine-tuning, T5-Base+MPOE on average improves by 2.13 for T5-Base due to shared central tensors. 
% Overall, our proposed approach obtains \textcolor{blue}{superior performance than Hyperformer~\cite{rabeeh2021hyperformer} }
% It further indicates the effectiveness of our proposed approach in multi-task learning.
It further demonstrates the potential benefits of the MPOE approach in a multi-task learning setting, where the central tensor learns common information across tasks and the auxiliary tensor learns task-specific information.
\begin{table}[t]
\centering
\small
\begin{tabular}{lrrr} 
\toprule
    \multicolumn{1}{c}{\multirow{1}{*}{Datasets}} & \multicolumn{1}{c}{T5-Base} & \multicolumn{1}{c}{Hyper$\clubsuit$} & \multicolumn{1}{c}{+MPOE} \\    \midrule
        MNLI~(acc)                &87.73  & 85.74 & \textbf{87.83}\\
        QNLI~(acc)               &93.51  & 93.02 & \textbf{93.89}\\
        SST-2~(acc)              &92.50  & 94.03 & \textbf{94.73}\\
        RTE~(acc)                 &75.41  & 75.36 & \textbf{75.51}\\
        QQP~(acc)               &91.12  & 90.28 & \textbf{91.17}\\
        CoLA~(mcc)               &54.93  & 63.73 & \textbf{65.85}\\
        MRPC~(acc)                &89.21  & 89.66 & \textbf{90.10}\\
        STS-B~(pearson)           &90.75  & 90.00 & \textbf{90.92}\\\midrule
        Avg.                     &84.39  & 85.23 & \textbf{86.25}\\ 
         {\#To~(M)}                   &223    & 343   & 258\\
\bottomrule
\end{tabular}
\caption{Performance of multi-task learning on GLUE benchmark obtained by fine-tuning T5-Base~(in percent). 
$\clubsuit$: Experimental results from Hyperformer~\cite{rabeeh2021hyperformer}.}
\label{tab:multi-task}
\end{table}

\subsection{Ablation Results}
% \begin{table}[t]
% \small
% \centering
% \begin{tabular}{lrrrr}
% \toprule
%   \multicolumn{1}{c}{\multirow{2}{*}{Variants}} &
%   \multicolumn{3}{c}{WikiText-2} &
%   \multicolumn{1}{c}{\multirow{2}{*}{\#To~(M)}} \\
%                      & PPL~$(\downarrow)$            & B2         & B4        & \\ \midrule
% w/o PS      & 21.28          & 28.67          & 9.44          & 
% 124.0\\
% w/o GM                  & 21.17          & 28.71           & 9.47         & 124.0\\
% w/o PS+GM     & 21.34          & 28.58          & 9.39          & 157.4\\ \midrule
% +MPOE                & \textbf{20.73} & \textbf{28.78} & \textbf{9.51}      & 157.4\\  
% \bottomrule 
% \end{tabular}
% \caption{Ablation study on the  WikiText-2 dataset about the NLG tasks~(in percent).
% ``B2'' and ``B4'' are short for BLEU-2 and BLEU-4, respectively.}
% \label{tab:ablation-results}
% \end{table}
\begin{table}[t]
\small
\centering
\begin{tabular}{lrrrr}
\toprule
  \multicolumn{1}{c}{\multirow{2}{*}{Variants}} &
  \multicolumn{3}{c}{WikiText-2} &
  \multicolumn{1}{c}{\multirow{2}{*}{\#To~(M)}} \\
                    & PPL~$(\downarrow)$            & B2         & B4        & \\ \midrule
+MoE$\bigstar$      & 21.86          & 28.27          & 9.14          & 578\\ \midrule
w/o PS              & 21.28          & 28.67          & 9.44          & 153\\
w/o GM              & 21.17          & 28.71           & 9.47         & 157\\ \midrule
+MPOE               & \textbf{20.72} & \textbf{28.78} & \textbf{9.51}      & 157\\  
\bottomrule 
\end{tabular}
\caption{Ablation study on the  WikiText-2 dataset about the NLG tasks~(in percent).
``B2'' and ``B4'' are short for BLEU-2 and BLEU-4, respectively. $\bigstar$: The method from~\citet{shazeer2017outrageously}}
\label{tab:ablation-results}
\end{table}

Our approach has incorporated two novel improvements: 
% 基于MPO分解的共享信息的混合专家模型，和用于快速训练的专家选择策略。
(1) MoE architecture with parameters sharing~(PS)  among experts based on MPO decomposition and (2) gradient mask~(GM) to alleviate unbalanced optimization.

To verify the effectiveness of each component, we conduct the ablation study on the WikiText-2 dataset to analyze the contribution of each part. We adopt PPL, BLEU-2 and BLEU-4 as the evaluation metrics, and consider removing the parameters sharing  and gradient mask strategy respectively.
The ablation results of our MPOE approach are shown in Table~\ref{tab:ablation-results}. We can see that removing any component would lead to a decrease in the model performance. It shows the effectiveness of all these components
in our approach. 
Besides, parameter sharing seems more important than the gradient mask strategy, which yields a larger performance drop after being removed. 
\subsection{Detailed Analysis}
\label{sec-detailed_analysis}
% MPO分解具有不同的分解方式。
MPO decomposition has different factorization manners.
However, the MPOE approach requires a defined MPO decomposition form to be given before it can be used.
Therefore, different factorization manners may affect the efficiency of the MPOE approach. 
To vertify this, we perform a detailed analysis on different factorization manners of MPO decomposition. 
We present three variants of MPOE with different lengths of local tensors produced by MPO decomposition empirically. 
% Tabel~\ref{tab-factorization-mannner} shows the results of our experiment on WikiText datasets, where $m=3,5,7$ corresponds to the number of local tensors.
Tabel~\ref{tab-factorization-mannner} shows the evaluation results on the WikiText-2 dataset about NLG tasks.
% As we can see, MPOE~(m=5) outperforms both the other variants and the GPT-2 model. 
% It indicates that the current design of our MPOE approach can efficiently utilize the limited data and alleviate the cold start problem. 
% As we can see, the MPOE variants for $m>3$ outperform GPT-2 model, and there are no significant differences among the variants. Considering the trade-off between the cost and quality, it is reasonable for us to finally choose the MPOE~(m=5).
As we can see, the variants of $m>3$ are all superior to the GPT-2 model. Additionally, we can observe that more local tensors performs similarly but leads to higher memory cost. Thus we finally choose to set $m=5$ for MPOE architecture considering the trade-off between the cost and quality.

\begin{table}[t]
\centering
\resizebox{\columnwidth}{!}{
    \begin{tabular}{lrrrr} 
    \toprule[1pt]
        \multicolumn{1}{c}{Variants} & 
        \begin{tabular}[c]{@{}c@{}} \\PPL~$(\downarrow)$ \end{tabular} &  \begin{tabular}[c]{@{}r@{}}WikiText-2\\B2\end{tabular} & \begin{tabular}[c]{@{}r@{}} \\B4\end{tabular} & 
        \begin{tabular}[c]{@{}r@{}}\#To~(M) \end{tabular} \\ 
        \midrule[0.7pt]
        GPT-2 & 21.27 & 28.69 & 9.46 & 124.4 \\
        MPOE~($m$=3) & 24.01 & 27.86 & 8.93 & 130.3 \\
        MPOE~($m$=5) & 20.72 & 28.77 & 9.48 & 157.4 \\
        MPOE~($m$=7) & 20.73 & 28.76 & 9.47 & 198.7 \\
        MPOE~($m$=9) & 20.78 & 28.45 & 9.38 & 214.6 \\
    \bottomrule[1pt]
\end{tabular}
}
\caption{Evaluation with different factorization manner on the  WikiText-2 dataset about the NLG tasks~(in percent).
``B2'' and ``B4'' are short for BLEU-2 and BLEU-4, respectively.}
\label{tab-factorization-mannner}
\end{table}
\section{Related Work}
We will review the related works in four aspects.

\paragraph{PLMs with MoE.}
% MoEshazer提出用于模型的扩容问题。
% Mixture-of-Experts architecture was introduced by~\citet{jacobs1991adaptive} to enlarge the model capacity.
% Pre-trained Language Models can achieve good performance in various NLP tasks by using a large number of parameters pre-trained on the general-purpose corpus~\citep{devlin2018bert,multitask2020raffel}. 
% 通常认为模型参数越多，模型的容量就越大。
It has been reported that models with more parameters are usually considered to have a larger model capacity~\citep{fedus2021switch,zuo2021thor}.
In order to increase the model capacity, a promising direction is to explore the scaling properties with MoE architecture which was introduced by~\citet{jacobs1991adaptive}.
Thus, \citet{shazeer2017outrageously} first applied the MoE architecture to  large-scale language models.
% Then, \citet{shazeer2017outrageously} first applied the MoE architecture to  large-scale language models.
% Since the MoE architecture was introduced~\citep{jacobs1991adaptive}, it has been the topic of numerous studies~\cite{collobert2002parallel,DBLP:conf/kdd/MaZYCHC18}.
Then, Switch Transformers~\citep{fedus2021switch}, GShard~\citep{dmitry2021gshard}, BASELayer~\citep{lewis2021base} and HashLayer~\citep{roller2021hash} studied how to build large-scale Transformer-based model with MoE as well as improving routing strategy, which can better utilize the model capacity.
In addition,~\citet{zhang2021moefication} proposed a strategy for sparse activation of MoE architecture.~\citet{he2021fastmoe} suggested a distributed training system for fast training of MoE.
~\citet{zoph2022designing} proposed a sparse expert model with more stable training.
~\citet{yu2022efficient} proposed a sparse expert model based on all-MLP architecture.
In contrast, our approach aims to reduce information redundancy by sharing parameters among experts.

\paragraph{Matrix Product Operators Decomposition.} 

Matrix product operators~(MPO)~\cite{pirvu2010matrix} decomposition was proposed in quantum many-body physics, \emph{a.k.a.} tensor-train~(TT) decomposition~\cite{oseledets2011tensor}.
A major category of MPO studies relies on model compression~\cite{gao2020compressing}. They focus on compressing weight matrix and convolutional layers~\cite{novikov2015tensorizing,garipov2016ultimate,sun2020model}.
Furthermore, the MPO decomposition was used to compress the PLMs as well as enable lightweight fine-tuning in downstream tasks~\cite{DBLP:conf/acl/LiuGZXLW20}.
In this work, we utilize such a decomposition mechanism for parameter sharing to construct a parameter-efficient MoE architecture.
% \textcolor{blue}{Our approach differs from these works in that we combine two compression ideas~(tensor representation and parameter sharing) to construct a tensorized parameter-efficient MoE architecture.}

% \textcolor{blue}{\paragraph{Improving deployment of MoE.}
\paragraph{Improved Variants of MoE.}
% 使用非常
Despite the achieved performance performance, MoE architecture has been hindered by the model complexity and high memory costs~\cite{shazeer2017outrageously, fedus2021switch}.
% Maintaining a large number of experts requires a very high memory consumption.
% This problem causes training a model based on the MoE architecture to become particularly expensive. 
This problem can be alleviated by using distillation~\cite{fedus2021switch} and expert pruning~\citep{kim2021scalable}.
% Similarly, ~\citet{kim2021scalable} proposed the use of pruning to improve the deployment of the model by removing experts that do not have an impact on the final results.
% Similarly, ~\citet{kim2021scalable} proposed the use of pruning to improve inference time efficiency by removing experts that do not have an impact on the final results.
Then, ~\citet{kudugunta2021beyond} and~\citet{zuo2021thor} indicated that sub-networks can be employed when using the model.
% Our approach can combine these methods to improve model deployment even further.
Indeed, our approach can be further enhanced by these existing methods for improving  inference time.

\paragraph{Multi-task Learning.}
The exploitation of MoE architectures for multi-task learning is a very promising direction in recent years~\citep{DBLP:conf/kdd/MaZYCHC18}.
~\citet{DBLP:conf/icml/HoulsbyGJMLGAG19} suggested training adapters for each task separately while keeping the model fixed. 
Further research suggested that model parameters could be shared across tasks, and task-specific adapter parameters were introduced~\cite{DBLP:conf/icml/Stickland019}. 
Based on this idea, ~\citet{rabeeh2021hyperformer} and ~\citet{pilault2020conditionally} proposed that parameter-efficient multi-task fine-tuning for transformer-based models via shared hypernetworks.
% \textcolor{blue}{Then, it was proposed that parameter-efficient multi-task fine-tuning for transformer-based models via shared hypernetworks~\citep{pilault2020conditionally,rabeeh2021hyperformer}.}
Our approach differs from these works in that the MPOE approach allows us to reduce model size while keeping the same number of experts, and meanwhile achieve performance improvement for multi-task learning.

\section{Conclusion}
In this paper, we proposed a parameter-efficient MoE architecture for increasing model capacity based on the MPO decomposition. 
First, we  shared the central tensors among different experts based on  MPO decomposition, which largely reduced the model parameters of MoE architecture.
Then, we designed the gradient mask strategy to alleviate the unbalanced optimization issues and ensured that different tensors capture different types of information efficiently.
% We also provided  both the theoretical analysis of the proposed MPOE approach and the existing improved methods for MoE architecture.
% \textcolor{blue}{Through MPO decomposition, we can effectively reorganize and aggregate information in the central tensors.
% Inspired by this, we design a novel, parameter-efficient MoE architecture by sharing parameters in central tensors among different experts.
% We also introduce the gradient mask strategy to alleviate unbalanced optimization and ensure that different tensors capture different types of information efficiently.}
Extensive experiments have shown that our approach outperforms several competitive PLM scaling strategies, especially in terms of improving the parameter efficiency of the MoE architecture. 
%To the best of our knowledge, this is the first application of MPO for increasing model size.

In the future, we will enhance the proposed MPOE approach with recently proposed routing methods, such as BASELayer~\cite{lewis2021base}, HASHLayer~\cite{roller2021hash} and GShard~\cite{dmitry2021gshard}. We will also consider exploring additional decomposition methods for developing parameter-efficient MoE architecture.

% \textcolor{blue}{In future work, we will consider exploring additional decomposition structures for MPO.}
% Note that MPOE still depends on Transformer architecture, and we will address this limitation in our future work.
% In future work, we will address the limitation that MPOE still depends on Transformer architecture.

\section*{Acknowledgments}
This work was partially supported by Beijing Natural Science Foundation under Grant No. 4222027,  National Natural Science Foundation of China under Grants No. 62206299 and 11934020, Beijing Outstanding Young Scientist Program under Grant No. BJJWZYJH012019100020098 and Beijing Academy of Artificial Intelligence (BAAI). Xin Zhao is the corresponding author.

%  the Fundamental Research Funds for the Central Universities and the Research Funds of Renmin University of China under Grant No. 18XNLG22, 19XNQ047, 20XNLG19 and 21XNH027

\bibliography{custom}
\bibliographystyle{acl_natbib}

% \appendix
% \input{appd}
% \section{Example Appendix}
% \label{sec:appendix}

% This is an appendix.

\end{document}